# ALPI: Auto-Labeller with Proxy Injection for 3D Object Detection using 2D Labels Only

Saad Lahlali    Nicolas Granger    Hervé Le Borgne    Quoc-Cuong Pham

Université Paris-Saclay, CEA, List, F-91120, Palaiseau, France

firstname.lastname@cea.fr

## Abstract

*3D object detection plays a crucial role in various applications such as autonomous vehicles, robotics and augmented reality. However, training 3D detectors requires a costly precise annotation, which is a hindrance to scaling annotation to large datasets. To address this challenge, we propose a weakly supervised 3D annotator that relies solely on 2D bounding box annotations from images, along with size priors. One major problem is that supervising a 3D detection model using only 2D boxes is not reliable due to ambiguities between different 3D poses and their identical 2D projection. We introduce a simple yet effective and generic solution: we build 3D proxy objects with annotations by construction and add them to the training dataset. Our method requires only size priors to adapt to new classes. To better align 2D supervision with 3D detection, our method ensures depth invariance with a novel expression of the 2D losses. Finally, to detect more challenging instances, our annotator follows an offline pseudo-labelling scheme which gradually improves its 3D pseudo-labels. Extensive experiments on the KITTI dataset demonstrate that our method not only performs on-par or above previous works on the Car category, but also achieves performance close to fully supervised methods on more challenging classes. We further demonstrate the effectiveness and robustness of our method by being the first to experiment on the more challenging nuScenes dataset. We additionally propose a setting where weak labels are obtained from a 2D detector pre-trained on MS-COCO instead of human annotations.The code is available at https://github.com/CEA-LIST/ALPI*

## 1. Introduction

In various fields such as autonomous vehicles, robotics, and augmented reality, achieving accurate 3D scene understanding stands as a critical task. This necessitates leveraging a diverse array of sensors, including cameras and LiDAR. Deep learning methods processing images [9, 15, 25] or point clouds [8, 20, 23] can extract valuable 3D spatial information. Their effectiveness largely depends on the availability of large-scale, diverse and precisely annotated data. For 3D object detection annotation, the process is nevertheless tedious, especially for occluded objects, with an average annotation time of 2 minutes per object [14].

In light of these challenges, weak supervision seeks to leverage annotations that do not fully correspond to the targeted task, but that are substantially cheaper to label. In the case of the 3D object detection task, instead of annotating the complete 3D box, a cheaper annotation can be the center of the 3D box [32], the center in bird-eye-view [14] or a 2D box annotated from a synchronized camera view [12, 13, 16, 27, 30]. These weakly supervised models are often used as 3D pseudo-labelers providing 3D box supervision for off-the-shelf fully supervised 3D detectors.

Weak supervision led to considerable progress in 2D-related tasks like 2D object detection [7, 28, 33], but remains at a relatively nascent stage for 3D-related tasks. Among existing 3D object detection approaches supervised with weak annotations (Table 1), one can identify two major limitations:

- Most of existing approaches [12–14, 16, 32] are actually semi-weakly supervised since they require a small amount of 3D box annotations;
- Approaches that do not require any 3D box annotation [27, 30] are specific to the class *Car* since they rely on heuristics that make them tedious or difficult to adapt to other classes.

We propose the first approach to address both limitations by being multi-class and requiring no 3D box annotation at all. More specifically, we supervise the model solely with readily available class size priors and 2D bounding boxes around objects in the camera views.

To be able to detect different object classes, we adopt a learning-based approach, leading to easier and better gen-eralization across various object classes. Nevertheless, the main challenge in this form of weak supervision stems from projection ambiguities: model predictions with 7 independent variables (size$\times$3, position$\times$3, and heading) are projected on images for comparison with 2D grountruths. Multiple 3D boxes can yield identical projected 2D boxes, leading to an ill-posed optimization problem. To address this issue, we propose the use of proxy classes which are simple cuboid-shaped objects sampled following size priors of real object classes. The synthetic proxy objects have 3D annotations by construction, allowing the model to learn 3D detection features that generalize to real object classes. Furthermore, to better handle hard examples, progressively replace proxy objects with real ones from confident pseudo-labels.

| Methods | Training annotations | | All object classes |
|---|---|---|---|
| | Min. required 3D boxes | Weak label type | |
| ViT-WSS3D [32] | 2% | 3D center | ✓ |
| WS3D [14] | (3.4%, 23.4%) | BEV center | ✓ |
| MTA [12], CAT [16], MAP-gen [13] | $\sim 13\%$ | 2D box | ✗ |
| FGR [27], GAL [30] | **0** | 2D box | ✗ |
| **ALPI** (ours) | **0** | 2D box | ✓ |

Table 1. Annotation settings for 3D object detection with weak annotations.

Our approach, named ALPI for Auto-Labeler with Proxy Injection, (i) is the first non-class-specific weakly supervised approach for 3D detection that does not require any 3D label. This contribution exhibits better performances on KITTI [4] *Cars* than previous comparable approaches. We also report unprecedented performance in this experimental setting (no 3D box label at all) on KITTI *Pedestrian* and *Cyclist*, as well as on nuScenes [24] to show the generalization ability of our approach. (ii) We propose a novel 2D loss to address range sensitivity; (iii) we further experiment replacing man-made 2D labels with predictions from a readily available 2D detector, opening the way to 3D detection on novel outdoor image and point cloud datasets *without any new human annotation*.

## 2. Related Work

Weak supervision yields substantial practical benefits when manual annotation is costly. In the context of 3D object detection, several annotation paradigms have been proposed.

**Semi-weakly supervised 3D object detection** methods adopt a mixed-annotation strategy where a small portion of the train set is 3D box annotated and the rest is weakly labeled. Different types of weak annotations have been proposed in this supervision setting. For instance, MTA [12], CAT [16] and MAP-Gen [13] adopt a semi-weak annotation strategy where 13% of the scenes are fully annotated while the rest are annotated with 2D bounding boxes. While these approaches achieve performance very close to fully supervised models, they focus on only one class and require a non-negligible amount of 3D box annotations. Methods working on more object classes and requiring fewer fully annotated scenes have been proposed but they require weak labels with 3D information. In particular, WS3D [14] used the object's center in bird's eye view and ViT-WSS3D [32] used the object's 3D center as weak annotation. While these weak annotations are appealing due to their valuable 3D information, it is important to note that they require annotating within the point cloud which increases the annotation time compared to 2D box annotation. As mentioned in [14], it is necessary to draw a 2D box on either the image or the point cloud to zoom in for easier and more precise point-annotation in a point cloud.

An original paradigm was proposed by Tang et al. [21]: a subset of object classes is fully annotated, while another distinct subset is weakly annotated with 2D boxes. The underlying concept involves extracting valuable 3D information from fully annotated classes to help detecting weakly annotated classes. By supervising weak classes with 2D boxes, the model learns their specific features. *Based on the empirical results of this method, we get the intuition that cross-class transfers depends more on size similarity than shape similarity. In a similar vein, our approach leverages cross-class transfer but from synthetic objects, thus requiring no manual annotations.*

**Weakly supervised 3D object detection** methods work under a more challenging setting where only weak labels are provided. Notably, FGR [27] and GAL [30] opted for 2D bounding boxes as weak annotations to detect *Cars* in 3D. Both methods rely on the standard pattern shape of *Cars* in bird's-eye view to handcraft heuristics. Therefore, designing new handcrafted heuristics for novel classes would be time-consuming, especially if these classes are not rigid (*Pedestrian*, *Cyclist*). *In opposition to these methods, we opt for a learning-based approach for easier and better generalization across diverse object classes.*

**Weak annotation paradigm**. Several paradigms have been proposed as they offer unique trade-offs between annotation costs and achieved performance. Approaches using few fully annotated scenes are motivated by the inherent challenges encountered when attempting to learn the detection of objects in 3D using only weakly annotated data. While these approaches outperform the ones which use no 3D box annotations, the selection of which scenes are fully annotated can significantly affect performance. This aspect was not investigated in semi-weakly supervised literature; however, Wang et al. [22] observed notable performance discrepancies when training a 3D detector with different small subsets of fully annotated scenes. *For this purpose,* *we set our method to be weakly supervised to lower dependency on the sampling of fully annotated scenes and opt for the cheaper 2D bounding box annotations as weak labels.*

## 3. Method

We adopt the typical weak supervision pipeline which trains a 3D annotator model in a weakly supervised fashion (Figure 2 part 2), then pseudo-annotates the dataset with 3D boxes (Figure 2 part 3) and finally trains an off-the shelf model [3, 18, 19] with these pseudo-labels (Figure 2 part 4).

**Working in a frustum point cloud.** By using the 2D box in an image and the point cloud, we can narrow down the search space for an object from the entire scene to a truncated pyramid (Figure 1) referred to as frustum.

### 3.1. Dataset Construction with Proxy Objects

Within this frustum, a majority of points usually belong to the object of interest while a small portion is held by the ground, occluding objects and the background. To better isolate the object from its background, the following step is to estimate its depth along the frustum axis.

**Size priors.** In our approach, size priors are used for various functions, including object depth estimation and background extraction. These priors were obtained from the internet by searching for information using only the class names. The same statistical data is consistently applied across all datasets. Detailed sources and statistical information (mean and standard deviation) are provided in the supplementary materials.

**Object depth estimation.** To determine the approximate depth of an object in a scene, we use the annotated 2D bounding box, the average object height $\tilde{H}$ and the LiDAR/camera projection matrix $P \in \mathbb{R}^{3\times4}$.

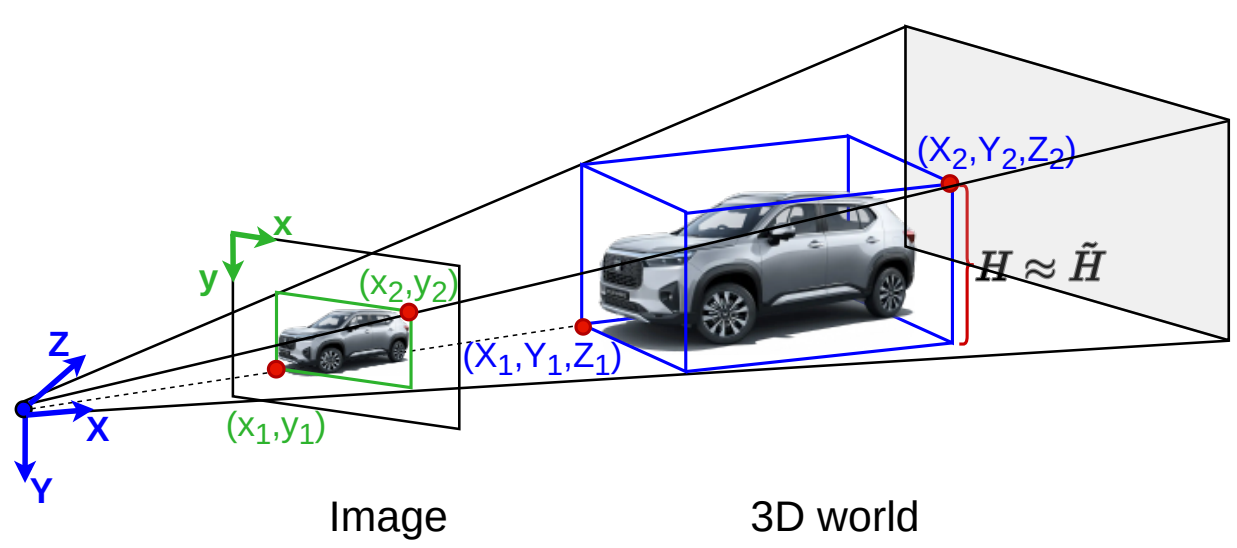

Figure 1. Illustration of the connection between the height of an object in the 3D world and the height of the projected 2D box annotation in the image. We take two points and their projection to estimate the object's depth.

The projection equation for the two points in red and purple in Figure 1 can be written as: $s_{1,2}[x, y_{1,2}, 1]^T = P[X, Y_{1,2}, Z, 1]^T$

By approximating the object's height as the average height i.e. $Y_2 - Y_1 = \tilde{H}$, we can estimate the depth of the object $X$ as:

$$\hat{X} = (K^{-1} s_2 \left[x,\ y_2,\ 1,\ 1/s_2\right]^T)_1 \tag{1}$$

$$\text{where } s_2 = \frac{\tilde{H}(P_{22} - \frac{y_1 P_{12}}{x})}{y_1 - y_2} \text{ and } K = \begin{bmatrix} P \\ 0\,0\,0\,1 \end{bmatrix}$$

**Background extraction.** We proceed to a coarse segmentation of the object within the frustum in order to extract the background. We design a simple approach which confidently filters out the target object points while potentially excluding some associated with occluding objects, the ground, or background. Points within the frustum are considered part of the background if they fall outside a cylinder centered at a depth $\hat{X}$ on the optical line and with a radius equal to the average length of the object $\tilde{L}$. This segmentation is not used for supervision but to filter out the object from the background scene which is used in the next part as context for object insertion.

**Object injection in the background.** To insert an object in the background scene, a 3D center position is needed. Since we want to substitute the previous object by a new one, we simply define the center as the median value w.r.t. each dimension of the points previously removed. The inserted object can either be a proxy or a pseudo-annotated object. The injection of pseudo-labeled objects is explained in subsection 3.3.

**Proxy objects.** A proxy object is designed with a cuboid-shaped appearance defined by a 3D center (as defined previously), a 3D size and a heading. The proxy's size is sampled using the size prior mean and standard deviation (part 1 in Figure 2). Among 12 possible headings in $[0, \pi)$, we select the one that best aligns the projection of the 3D box with the previous 2D box annotation. After positioning the proxy in the scene, points are randomly sampled on the surfaces visible to the LiDAR (Figure 3).

### 3.2. 3D Annotator Architecture and Training

**Architecture.** We opt for Frustum-ConvNet [26] as architecture for the 3D annotator. This model estimates the 3D box of an object within a frustum. At each training iteration, two types of frustums are fed to the 3D annotator. As shown in Figure 2, in green the first type being frustums with target objects (*Cars, Pedestrians* and *Cyclists*) which are supervised in 2D only. The second type, in blue, are frustums with injected object (proxies or pseudo-labeled) which are supervised both in 2D and 3D. The annotator model predicts a $Box_{3D}$ which encompasses the 3D coordinates of the center, sizes, and heading of the object. The 8 corners

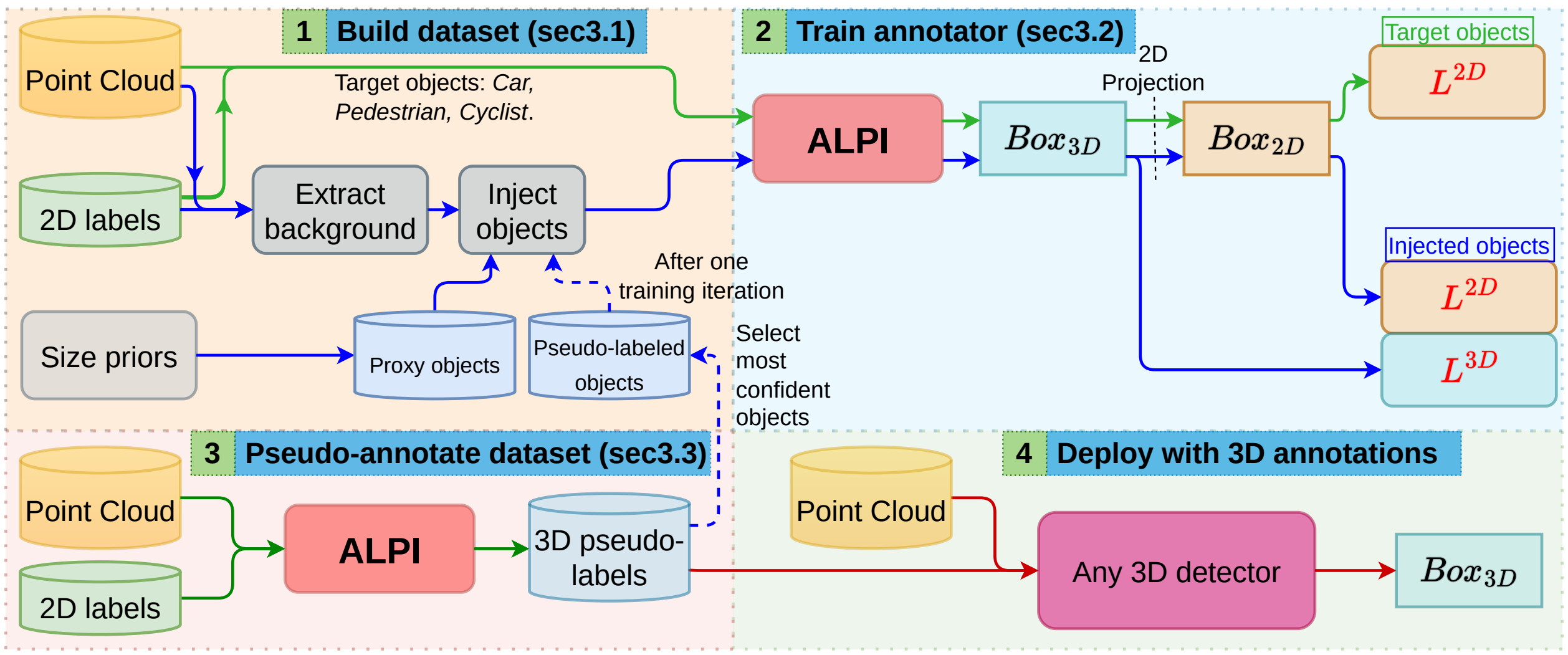

Figure 2. **Overview of our approach.** Our contribution consists in a 3D annotator **ALPI** that provides the 3D pseudo-labels, using only point cloud data and weak annotations (2D boxes) as input (part 3 in Figure 2). This 3D annotator is learned (part 2 of Figure 2) using not only point cloud data and weak annotation but also *proxy objects* that are injected in point cloud data. From our intuition about what makes cross-class transfer fruitful, we propose to build proxy objects using size priors only and to inject them in the scenes (part 1 of Figure 2). Our 3D annotator learns to estimate 3D boxes for both target and injected objects. It is trained by supervising targets in 2D and injections in 3D. Finally, an iterative refinement scheme gradually improves the quality of 3D pseudo-labels in order to retrieve harder examples.

of this $Box_{3D}$ are projected in 2D and the minimum and maximum w.r.t each dimension produces the $Box_{2D}$. We minimize a multi-task loss that is the sum of several terms:

**Classification loss.** We apply a focal loss function [10] to the class predictions of the model. Such loss has already been used by [32] for semi-weakly supervised 3D object detection.

**Losses on $\mathbf{Box_{3D}}$.** When the frustums contain proxy objects or pseudo-labelled objects that provide a 3D pseudo ground-truth, we apply regression losses on the estimated $Box_{3D}$ of these instances. The 3D center coordinates are regressed with the Huber loss [6], while the size is supervised via a smooth L1 loss. To handle the range of object headings within the interval $[0, \pi)$, we discretize it into 12 bins and pose the problem of estimating the heading as a classification task, with a cross-entropy loss. The residual heading offset is supervised through a smooth L1 loss.

**Proposed Loss on $\mathbf{Box_{2D}}$.** For both target and injected objects, we supervise the projection of the predicted 3D box in the image ($Box_{2D}$) using available 2D bounding box labels. Typically, a 2D detection regression loss would use a smooth L1 loss ($\ell$) on each $Box_{2D}$ coordinate, giving a total loss $L^{2D} = \ell(\hat{x}_1, x_1) + \ell(\hat{x}_2, x_2) + \ell(\hat{y}_1, y_1) + \ell(\hat{y}_2, y_2)$, where $x_., y_.$ and $\hat{x}_., \hat{y}_.$ are the coordinates of the annotated and predicted 2D boxes, respectively. However, when a predicted $Box_{3D}$ is shifted from its ground truth by a constant distance and translated across different depths, the 3D IoU error remains constant, but the 2D projection causes the $L_{2D}$ loss to vary with object depth. This is problematic in outdoor scenes with diverse object depths (0 to 70 meters), as changing depth distributions in training batches can cause large variations in mean loss, disrupting smooth training. To address this, we propose a depth-normalized 2D loss, leveraging the relationship between the 2D box size and object depth discussed in section 3.1. The new loss is:

$$L^{2D}_{norm} = \frac{\ell(\hat{x}_1, x_1) + \ell(\hat{x}_2, x_2)}{x_2 - x_1} + \frac{\ell(\hat{y}_1, y_1) + \ell(\hat{y}_2, y_2)}{y_2 - y_1} \tag{2}$$

This formulation normalizes the 2D loss by the height and width of the ground-truth 2D bounding box, ensuring consistent loss values regardless of object depth.

**Size regularization.** We apply a regularization loss on the size of the 3D boxes predicted by the model at each training batch computed as:

$$L^{3D}_{regu} = \sum_{k \in \{h,l,w\}} \ell(\hat{\sigma}_k, \sigma_k) + \ell(\hat{\nu}_k, \nu_k) \tag{3}$$

where $\sigma_., \nu_.$ and $\hat{\sigma}_., \hat{\nu}_.$ denote the standard deviation and average sizes in a training batch and in the validation set respectively, $\ell$ is a smooth L1 loss and $h$, $l$, $w$ are respectively the object's height, length and width.

### 3.3. 3D Pseudo-Annotation of the Dataset

The 3D annotator is used to annotate the dataset in two cases. First, in order to pseudo-label objects before injecting them in the background scene. Secondly, for training a 3D detector during deployment. To improve its robustness in detecting occluded samples, we inject confident pseudo-labeled objects into background scenes and iteratively retrain the annotator.

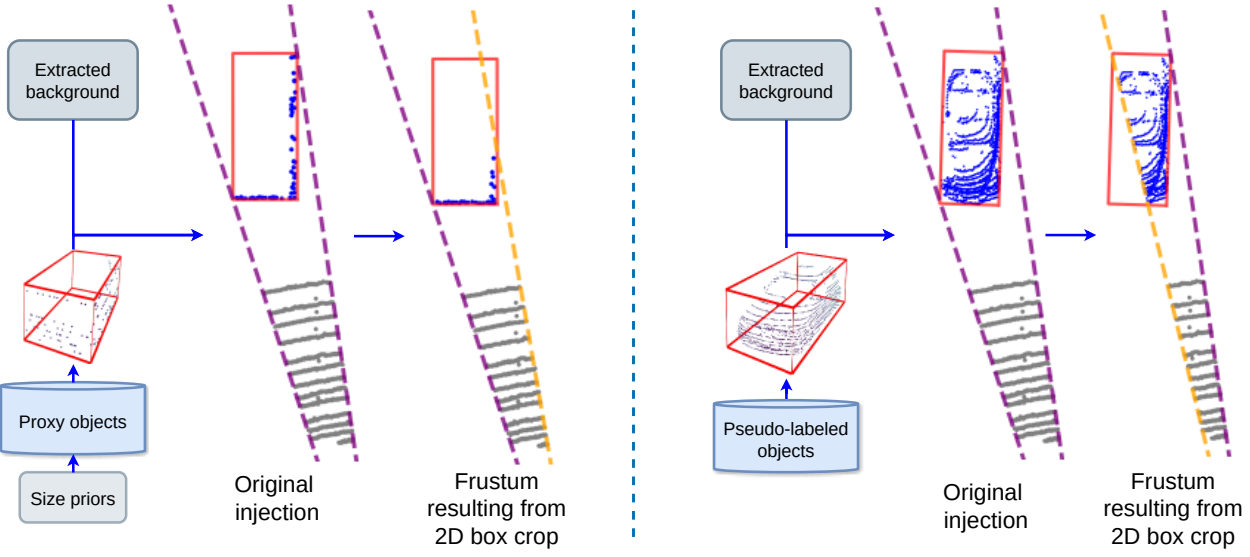

Figure 3. Injection of proxy objects on the left and of pseudo-labeled objects on the right. To achieve better robustness to occluded objects we crop the 2D box annotation and build from this crop a new frustum which is tighter. The original frustum lines are shown in purple and after cropping are shown in orange. The objects points are colored in blue and background in grey. For easier visualisation, we present only a zoom-in of the frustum.

At iteration 0, the 3D annotator trains on target and proxy objects only. After each iteration, we save a catalog of pseudo-annotated objects, defined by the predicted 3D box and the points within this box (Figure 3). We trust only pseudo-labels where the projected $Box_{2D}$ closely matches the ground-truth 2D box, indicating likely correct predictions. These trusted pseudo-labeled instances are injected into background scenes later. We select pseudo-labeled objects with an azimuth angle close to that of the removed object, referring to the angle between the ego *Car* and the object's center in a bird's eye view. The pseudo-labeled object is then translated along the azimuth to its new 3D position (section 3.1). To enhance model robustness to occlusions, the injected object is augmented (Figure 3).

While previous studies have explored augmentation with object injections [5, 29], we specifically design the injection of pseudo-labeled objects to improve robustness for cases with imperfect 2D box annotations (Figure 3). We simulate a cropped version of the 2D box from the 3D box's 2D projection and use this cropped 2D box to extract the frustum, integrating objects with accurate 3D box estimations and reproducing the appearance of occluded objects in the point cloud.

### 3.4. Towards Zero New Human Annotation

To reduce annotation costs, we propose using 2D boxes generated by a pretrained 2D detector instead of an Oracle capable of annotating complete 2D boxes. We use the Faster R-CNN detector [17], pre-trained on MS-COCO [11] from the MMDet toolbox [1], to generate these weak labels.

Compared to Oracle 2D boxes from KITTI, typical 2D detector boxes outline only the visible portion of objects, not the full 3D projection. This difference is particularly significant in occluded scenarios, as shown in Figure 4, where the extrapolated 2D box is green, and the Faster R-CNN box is purple. Consequently, incomplete weak annotations of occluded objects cannot directly train the 3D annotator, necessitating method adaptation.

At iteration 0, we refine the training set by excluding occluded objects with incomplete 2D boxes, using a 2D detector confidence threshold of 0.95. The 3D annotator then pseudo-labels the entire dataset. In subsequent iterations, we inject these pseudo-labeled objects into background scenes, filtered by a 0.95 confidence score from the 2D detector. Instead of weak annotations from the 2D detector, we use the more accurate 2D projections inferred by the 3D annotator (orange frustum lines in Figure 4).

The frustum-based architecture [26] allows selective exclusion of frustums with low-confidence 2D box annotations, simplifying the filtering process compared to more complex scene-based solutions.

## 4. Experiments

### 4.1. Dataset and Evaluation Metric

We validate our approach on the KITTI dataset [4], which provides 7,481 pairs of RGB images and point clouds for training and 7,518 pairs for testing. The dataset includes annotations for three object categories (*Car*, *Pedestrian*, and *Cyclist*), each evaluated under three difficulty levels (easy, moderate, and hard) based on object occlusion and truncation. As the test set ground truth is hidden, we follow existing works [2] and split the training set into 3,712 training and 3,769 validation instances.
We evaluate 3D detection performance using mean Average Precision (mAP). mAP measures the accuracy of object localization and classification by averaging precision across different classes and IoU thresholds, reflecting a model's ability to detect and localize objects in 3D space accurately. We compute mAP at 40 recall points with a 3D IoU threshold of 0.7 for *Car* and 0.5 for *Pedestrian* and *Cyclist*.

### 4.2. Implementation details.

The training of the 3D annotator follows the settings set by the original F-ConvNet model [26]. We changed the Adam optimizer to a weight decay of 0.0001 and a learning rate that starts from 1e-4 and decays by a factor of 10 every 20 epochs for a total of 50 epochs. We set a classic depth range of [0, 70] meters in KITTI. For the proxy objects, we randomly sample between 50 and 200 points on 6 different horizontal lines. We consider the 3D annotator to have converged after 2 pseudo-labeling iterations. At iteration 0, 1 and 2, the proportions of pseudo-labeled instances over proxy objects during injection are respectively 0, 0.3 and 0.3. After iteration 2, we apply the same refinement as in the original F-ConvNet model and use the whole pseudo-labeled train set for training.

| 3D detector | Paradigm | 3D annotator | **Cars** IoU@0.7 | | | **Ped.** IoU@0.5 | | | **Cyclist** IoU@0.5 | | |
|---|---|---|---|---|---|---|---|---|---|---|---|
| | | | easy | mod. | hard | easy | mod. | hard | easy | mod. | hard |
| Point-RCNN [19] | fully | *Oracle* | 89.75 | 80.35 | 77.94 | 61.50 | 52.95 | 46.36 | 90.93 | 71.40 | 66.98 |
| | semi-weak | ViT-WSS3D [32] | 83.5 | 72.7 | 70.1 | 57.1 | 51.9 | 46.8 | 72.8 | 55.0 | 51.6 |
| | weak | FGR [27] | 86.40 | 73.87 | 67.03 | - | - | - | - | - | - |
| | | GAL [30] | 88.21 | 76.13 | 69.61 | - | - | - | - | - | - |
| | | **ALPI (ours)** | **90.48** | **79.43** | **75.81** | **59.79** | **50.97** | **45.09** | **87.80** | **68.06** | **61.43** |
| PV-RCNN [18] | fully | *Oracle* | 92.57 | 84.83 | 82.69 | 68.06 | 60.80 | 56.30 | 89.76 | 72.90 | 68.32 |
| | semi-weak | ViT-WSS3D | 84.5 | 75.8 | 71.1 | 61.0 | 53.4 | 48.2 | 67.4 | 49.7 | 46.2 |
| | weak | FGR [27] | 83.79 | 67.77 | 65.68 | - | - | - | - | - | - |
| | | GAL [30] | 84.03 | 67.83 | 66.10 | - | - | - | - | - | - |
| | | **ALPI (ours)** | **90.49** | **81.09** | **74.79** | **64.07** | **55.78** | **53.84** | **84.97** | **66.28** | **63.94** |
| Voxel-RCNN [23] | fully | *Oracle* | 92.38 | 85.29 | 82.86 | - | - | - | - | - | - |
| | weak | FGR [27] | 84.56 | 68.29 | 66.23 | - | - | - | - | - | - |
| | | GAL [30] | 85.98 | 69.36 | 66.71 | - | - | - | - | - | - |
| | | **ALPI (ours)** | **91.16** | **81.43** | **76.72** | - | - | - | - | - | - |

Table 2. 3D object detection AP (%) on KITTI **val** set. Best per paradigm in **bold**. For each 3D detector, the upper-bound in gray corresponds to the fully-supervised setting.

| 3D detector | Paradigm | 3D annotator | **Cars** IoU@0.7 | | | **Ped.** IoU@0.5 | | | **Cyclist** IoU@0.5 | | |
|---|---|---|---|---|---|---|---|---|---|---|---|
| | | | easy | mod. | hard | easy | mod. | hard | easy | mod. | hard |
| Point-RCNN [18] | fully | *Oracle* | 85.94 | 75.76 | 68.32 | 49.43 | 41.78 | 38.63 | 73.93 | 59.60 | 53.59 |
| | weak | FGR [27] | 80.26 | 68.47 | 61.57 | - | - | - | - | - | - |
| | | GAL [30] | **82.73** | 70.02 | 62.33 | - | - | - | - | - | - |
| | | **ALPI (ours)** | 80.40 | **72.46** | **68.05** | **44.21** | **36.42** | **32.82** | **72.18** | **55.59** | **49.92** |

Table 3. 3D object detection mAP (%) on KITTI **test** set when using Point-RCNN [19] as the 3D detector. Best for the weak paradigm in **bold**. Second best for the weak paradigm underlined. Due to the handcrafted design of FGR [27] and GAL [30] on the class *Car* of KITTI, their adaptation to the classes *Pedestrian* and *Cyclist* is not possible.

### 4.3. Comparisons with the State of the Art

Following previous methods [27,30,32], we measure the effectiveness of our 3D annotator by training different off-the-shelf 3D detectors (PointRCNN [19], PV-RCNN [18], Voxel-RCNN [23]) with the obtained annotated 3D box pseudo-labels.

In Table 2, we report the results on KITTI validation set. Compared to previous weakly supervised methods, pseudo-labels generated by our method significantly improve the performance of all three 3D detectors. For instance, for the class *Car* and under the *moderate* difficulty, our method improves the performance of PointRCNN, PV-RCNN and Voxel-RCNN by 3.3%, 13.3% and 12.1% respectively. Moreover, with our 3D annotator, the 3D detectors have a more consistent gap of performance compared to their fully supervised setting. It has to be noted that, while PV-RCNN and Voxel-RCNN achieve better performances than PointRCNN under the fully supervised setting, with previous weakly supervised 3D annotators, this order changes by a large margin. For the class *Car* and under the *moderate* difficulty, a fully supervised PV-RCNN has a mAP 4.5% higher than a fully supervised PointRCNN, however when trained annotations from GAL [30] or FGR [27] as 3D annotator, the order is reversed and PointRCNN now has a mAP 8.3% and 6.1% higher than PV-RCNN, respectively. Under the same class and difficulty setting, using the annotations produced by our method, the order is conserved. Compared to the fully supervised setting, for PRCNN, PVRCNN and VoxelRCNN, the mAP gap is 0.9%, 3.7% and 3.9%, respectively.

To verify the generality of our 3D annotator, we evaluate on classes *Pedestrian* and *Cyclist* and compare its performances to ViT-WSS3D [32] which is under the semi-weak setting where 2% of the scenes are fully annotated. We notice that our method achieves better performances on architectures PointRCNN and PV-RCNN and across all classes. For example, on the *moderate* difficulty and on classes *Pedestrian* and *Cyclist*, PV-RCNN gains 2.38% and 16.58% in mAP respectively.

The results on KITTI test set (Table 3) are obtained by server submission. For fair comparison with previous methods [27, 30], we use a per-class PointRCNN as 3D detector even though our method achieves better performances on the val set with other detectors. Our method outperforms previous works under the weak annotation setting by 2.4% in mAP on *moderate* and 5.7% on *hard* difficulty but not on the *easy* one. The gap in mAP performance comparatively to a fully supervised PointRCNN on *moderate Car*, *Pedestrian* and *Cyclist* is 3.3%, 5.36% and 4.0%.

### 4.4. Evaluation on nuScenes dataset

To demonstrate the broad applicability of our method across various object classes and scenarios, we conducted experiments on the nuScenes dataset under the weak label setting, being the first to do so, and present the results in Table 4. We employed CenterPoint [31] as the 3D detector model, trained with the 3D pseudo-labels generated by our 3D annotator. Since no weak or semi-weak approaches have reported performance on nuScenes, we compared our method with the fully supervised CenterPoint [31] and the state-of-the-art semi-supervised approach [24] which also uses CenterPoint. The nuScenes dataset is more challenging than KITTI due to a higher number of objects per scene and sparser point clouds, resulting in a larger performance gap between a 3D detector trained with oracle 3D annotations and one trained with our pseudo-annotations. In a true weakly supervised fashion, *we decided to leave all hyper-parameters of our method unmodified*. Despite not using temporal consistency, which significantly enhances performance [24], our method outperforms the semi-supervised approach on average. For a few classes like trucks with ambiguous shapes, our method performs lower, likely due to the resemblance of such objects to two connected 3D boxes. These findings aim to open evaluation for weakly supervised 3D object detection across a wider range of classes and datasets.

### 4.5. Ablations studies

**Role of proxy objects.** In the absence of proxy objects, the model can't converge well during training due to the ambiguities left by the 2D supervision alone. Proxy objects serve to initiate and enhance the model's detection capabilities, effectively transferring to real object classes to the extent that inferred pseudo-labels become viable for subsequent training iterations.

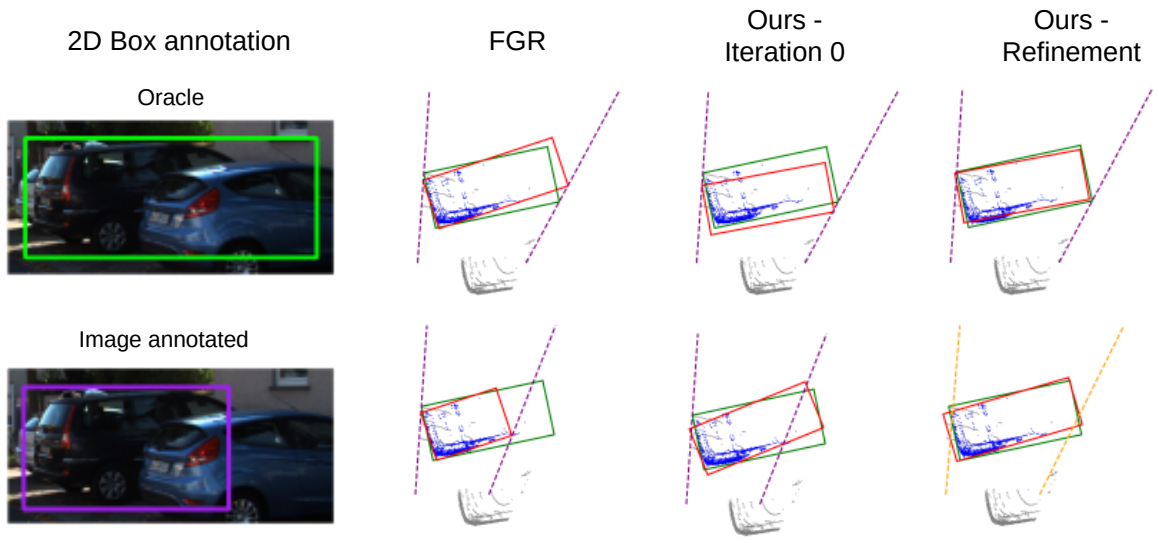

Figure 4. Comparison of annotations produced by our method and FGR. 2D box weak labels are either provided by an Oracle which refers to a complete 2D box (as used in FGR [27] and GAL [30]) or annotated on the image which refers to annotating only visible parts of the object (here provided by Faster R-CNN [17]). Predicted 3D boxes are drawn in red while ground truth boxes are in green. The 2D box annotations used during training are shown in the left. For better visualization, we color in blue the points belonging to the object of interest, in gray occluding object and in purple the frustum lines drawn using 2D box annotations. Additional qualitative results can be found in the supplementary materials.

**Effect of the depth normalized 2D bounding box loss.** The loss $L^{2D}_{norm}$ defined by Equation 2 takes into account how far the object is from the LiDAR sensor. Therefore, to investigate how much it increases the performances, we propose to experiment on the KITTI validation set using a usual smooth L1 loss $L^{2D}$ and compare it with our normalized $L^{2D}_{norm}$. The $L^{2D}_{norm}$ improves detection performances on the class Car in Table 5 and also on the other classes in the supplementary material, both with and without the regularization $L^{3D}_{regu}$ defined by Equation 3.

**Effect of pseudo-labeling iterations.** In Table 6 (with manual weak annotations), we report the performances of our method on the KITTI validation set after each of the pseudo-labeling iterations. It shows that the pseudo-labeling improves the model's performance after each iteration. Visualisation are provided in Figure 4 and Figure 5. Additional quantitative results can be found in the supplementary materials.

**Weak labels from any automatic 2D detector** In Table 6 (automatic weak annotation), we demonstrate the impact of pseudo-labeling iterations using imperfect weak labels obtained from Faster R-CNN (subsection 3.4). For the *moderate* difficulty, performance doubled from iteration 0 to the last, reaching a mAP of 70%, compared to a 14% increase to 76% with manual weak annotations. Additionally, our 3D annotator struggles more with harder examples compared to using oracle weak annotations, consistent with the fact that hard examples are often occluded. The bounding box predicted by Faster R-CNN, which encloses only the visible part of the object, differs from the extrapolated box containing the whole object. Unlike pure geometrical approaches like GAL [27] and FGR [27], where the 3D box boundaries are tied to the 2D weak label, our method, based on learned object representations, is better at recovering from noisy labels, especially in occluded situations (see Figure 4). FGR's pipeline, due to extensive filtering, cannot effectively use automatic 2D weak labels, resulting in partial scene annotations unsuitable for training 3D detectors.

| Paradigm | 3D boxes | 3D Annotator | mAP | Car | Truck | C.V. | Bus | Trailer | Barrier | Motor. | Bike | Ped. | T.C. |
|---|---|---|---|---|---|---|---|---|---|---|---|---|---|
| fully | 100% | *Oracle* | 58.8 | 84.8 | 57.5 | 18.3 | 69.3 | 34.8 | 68.5 | 57.2 | 42.8 | 85.3 | 69.9 |
| semi | 7% | [24] | 36.5 | 74.9 | 29.9 | 3.8 | 31.3 | 10.1 | 45.6 | 31.9 | 12.3 | 75.6 | 49.1 |
| weak | 0% | FGR [27] | - | - | - | - | - | - | - | - | - | - | - |
| | | GAL [30] | - | - | - | - | - | - | - | - | - | - | - |
| | | **ALPI (ours)** | 41.8 | 81.0 | 33.9 | 7.7 | 45.3 | 30.0 | 40.0 | 42.7 | 26.3 | 73.9 | 37.0 |

Table 4. 3D object detection performances in mAP on the **nuScenes** validation set. 'C.V.', 'Ped.', and 'T.C.' are short for 'construction vehicle', 'pedestrian', and 'traffic cone', respectively. **CenterPoint** [31] is used as the 3D detector. As no weak or semi-weak methods have reported on nuScenes, we compared our results with a fully and a semi-supervised approach [24]. nuScenes being more challenging than KITTI, the gap with the fully supervised is wider. On average across all classes, our method outperforms the semi-supervised approach.

| $L^{2D}$ | $L^{2D}_{norm}$ | $L^{3D}_{regu}$ | **Cars** IoU@0.7 | | |
|---|---|---|---|---|---|
| | | | easy | mod. | hard |
| ✓ | | | 78.23 | 69.71 | 59.08 |
| ✓ | | ✓ | 84.54 | 72.81 | 62.77 |
| | ✓ | | 89.96 | 78.02 | 72.53 |
| | ✓ | ✓ | **90.48** | **79.43** | **75.81** |

Table 5. 3D object detection mAP (%) on KITTI **val** set with a per-class PointRCNN trained with ALPI's 3D annotations.

| **Successive steps** | | easy | mod. | hard |
|---|---|---|---|---|
| **ALPI** | Iteration 0 | 68.94 | 64.81 | 56.73 |
| | Iteration 1 | 87.10 | 76.49 | 74.67 |
| | Iteration 2 | 87.84 | 77.12 | 76.48 |
| | Refinement | 88.97 | 77.80 | 77.48 |
| PointRCNN [19] trained with **ALPI**'s pseudo-labels | | **90.48** | **79.43** | **75.81** |

Table 6. 3D object detection mAP (%) on KITTI **val** set for the class Car. 3D pseudo-labels generated by ALPI at the refinement step are used to train the 3D detector PointRCNN.

| **Weak label source** (2D boxes) | easy | mod. | hard |
|---|---|---|---|
| Faster R-CNN | 83.66 | 74.64 | 67.16 |
| Oracle | 90.48 | 79.43 | 75.81 |

Table 7. 3D object detection mAP (%) on KITTI **val** set, with weak annotation (2D boxes) obtained manually (*Oracle*) and automatically with Faster R-CNN [17].

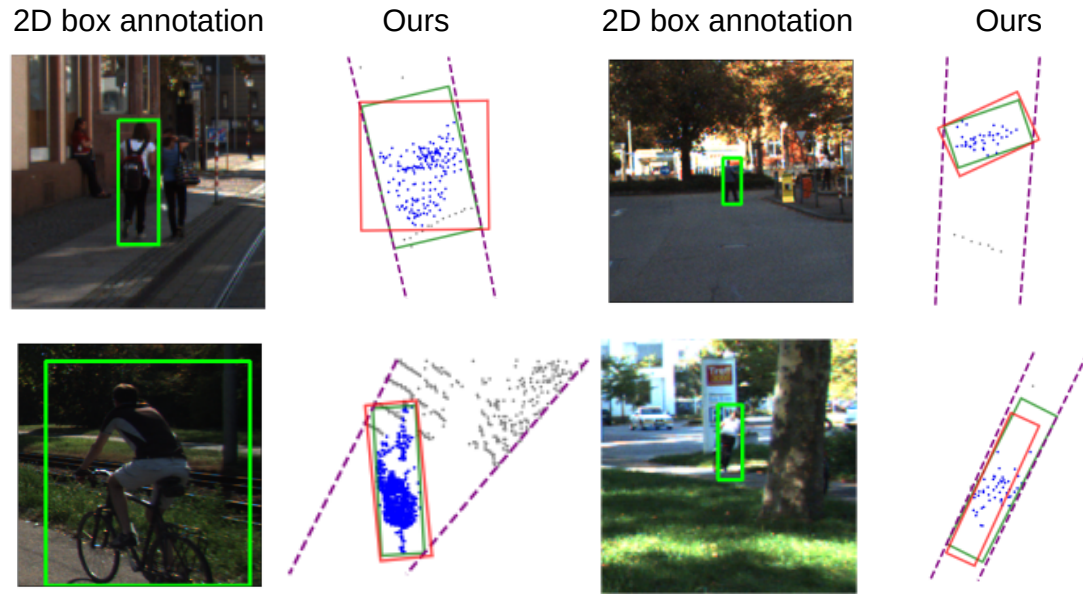

Figure 5. Predictions on classes *Pedestrian* and *Cyclist*. Predicted 3D boxes are drawn in red while ground truth boxes are in green. For better visualisation, we color in blue the points belonging to the object of interest and in grey the background. The 2D box annotations used during training are shown in the left. In purple are shown the frustum lines drawn using 2D box annotations.

## 5. Conclusions, Limitations and Perspectives

In the vein of Weakly Supervised 3D Vehicle Detection approaches [27, 30], the method we propose does not need any 3D annotation, contrary to most works in the literature that are Semi-weakly supervised 3D object detection methods. These two previous works were nevertheless limited to the *Car* class, while our approach can easily adapt to other ones (*Pedestrian*, *Cyclist*) simply by adding a corresponding proxy object class during training. The results on nuScenes motivate us to explore in future works methods which leverage temporal consistency to improve detection.

Being frustum-based, our method is better suited to large-scale scenes (ex: outdoor contexts), thus we will investigate to adapt it in order to relax its dependency to the frustum model.

This work was supported by a French government grant managed by the Agence Nationale de la Recherche under the France 2030 program with the reference "ANR-23-DEGR-0001". This publication was made possible by the use of the FactoryIA supercomputer, financially supported by the Ile-De-France Regional Council.